\documentclass[11pt,letterpaper]{article}
\usepackage{naaclhlt2013}
\usepackage{times}
\usepackage{latexsym}
\usepackage{hyperref}
\title{One Billion Word Benchmark for Measuring Progress in\\
Statistical Language Modeling}

\author{Ciprian Chelba, Tomas Mikolov, Mike Schuster, Qi Ge, Thorsten Brants\\
  Google\\
  1600 Amphitheatre Parkway\\
  Mountain View, CA 94043, USA\\
  \AND Phillipp Koehn\\
  University of Edinburgh\\
  10 Crichton Street, Room 4.19\\
  Edinburgh, EH8 9AB, UK\\
  \And Tony Robinson\\
  Cantab Research Ltd\\
  St Johns Innovation Centre\\
  Cowley Road, Cambridge, CB4 0WS, UK\\
}

\begin{document}
\maketitle
\begin{abstract}
We propose a new benchmark corpus to be used for measuring progress in statistical language modeling.
With almost one billion words of training data, we hope this benchmark will be useful
to quickly evaluate novel language modeling techniques, and to compare their contribution
when combined with other advanced techniques. We show performance of several well-known
types of language models, with the best results achieved with a recurrent neural network
based language model. The baseline unpruned Kneser-Ney 5-gram model achieves perplexity 67.6.
A combination of techniques leads to 35\% reduction in perplexity, or 10\% reduction in
cross-entropy (bits), over that baseline.

The benchmark is available as a code.google.com project; besides the scripts needed to
rebuild the training/held-out data, it also makes available log-probability values for
each word in each of ten held-out data sets, for each of the baseline n-gram models.

\end{abstract}

\section{Introduction}

Statistical language modeling has been applied to a wide range of applications and domains with great success. 
To name a few, automatic speech recognition, machine translation, spelling correction, 
touch-screen ``soft'' keyboards and many natural language processing applications depend on the quality of 
language models (LMs).

The performance of LMs is determined mostly by several factors: the amount of
training data, quality and match of the training data to the test data, and choice of modeling technique for estimation from the data. It is widely accepted that the amount of data, and the ability of a given estimation algorithm to accomodate large amounts of training are very important in providing a solution that competes successfully with the entrenched n-gram LMs. At the same time, scaling up a novel algorithm to a large amount of data involves a large amount of work, and provides a significant barrier to entry for new modeling techniques. By choosing one billion words as the amount of training data we hope to strike a balance between the relevance of the benchmark in the world of abundant data, and the ease with which any researcher can evaluate a given modeling approach. This follows the work of Goodman~\shortcite{Goodman:2001a}, who explored performance of various language modeling techniques when applied to large data sets. One of the key contributions of our work is that the experiments presented in this paper can be reproduced by virtually anybody with an interest in LM, as we use a data set that is freely available on the web.

Another contribution is that we provide strong baseline results with the currently very popular
neural network LM~\cite{Bengio:2003}. This should allow researchers who work
on competitive techniques to quickly compare their results to the current state of the art.

The paper is organized as follows: Section 2 describes how the training data was obtained; Section 3 provides a short overview of the language modeling techniques evaluated; finally, Section 4 presents results obtained and Section 5 concludes the paper.

\begin{table*}
\begin{center}
\begin{tabular}{|l|c|c|c|c|}
\hline
\bf Model                           & \bf Num. Params & \multicolumn{2}{c|}{\bf Training Time} & \bf Perplexity \\
\bf                                 & \bf [billions]  & \bf [hours]            & \bf [CPUs] &    \\
\hline
Interpolated KN 5-gram, 1.1B n-grams (KN) & 1.76      &   ~3                   & ~100 &           67.6 \\
Katz 5-gram, 1.1B n-grams           & 1.74            &   ~2                   & ~100 &           79.9 \\
Stupid Backoff 5-gram (SBO)         & 1.13            &   ~0.4                 & ~200 &           87.9 \\
\hline
Interpolated KN 5-gram, 15M n-grams & 0.03            &   ~3                   & ~100 &          243.2 \\
Katz 5-gram, 15M n-grams            & 0.03            &   ~2                   & ~100 &          127.5 \\
\hline
Binary MaxEnt 5-gram (n-gram features) & 1.13         &   ~1                   &  5000 &         115.4 \\
Binary MaxEnt 5-gram (n-gram + skip-1 features) & 1.8 &   1.25                 &  5000 &         107.1 \\
Hierarchical Softmax MaxEnt 4-gram (HME) &  6         &   3                    &  1    &         101.3 \\
\hline
Recurrent NN-256 + MaxEnt 9-gram    & 20              &   ~60                  &  24   &          58.3 \\
Recurrent NN-512 + MaxEnt 9-gram    & 20              &   ~120                 &  24   &          54.5 \\
Recurrent NN-1024 + MaxEnt 9-gram   & 20              &   ~240                 &  24   &          51.3 \\
\hline
\end{tabular}
\end{center}
\caption{\label{results-table}Results on the 1B Word Benchmark test set with various types of language models.}
\end{table*}

\section{Description of the Benchmark Data}

In the following experiments, we used text data obtained from the WMT11 website\footnote{\url{http://statmt.org/wmt11/training-monolingual.tgz}}. The data preparation process was performed as follows:
\begin{itemize}\addtolength{\itemsep}{-0.5\baselineskip}
\item All training monolingual/English corpora were selected
\item Normalization and tokenization was performed using scripts distributed from the WMT11 site, slightly augmented to normalize various UTF-8 variants for common punctuation, e.g. \verb+'+
\item Duplicate sentences were removed, dropping the number of words from about 2.9 billion to about 0.8 billion (829250940, more exactly, counting sentence boundary markers \verb+<S>+, \verb+<\S>+)
\item Vocabulary (793471 words including sentence boundary markers \verb+<S>+, \verb+<\S>+) was constructed by discarding all words with count below 3
\item Words outside of the vocabulary were mapped to \verb+<UNK>+ token, also part of the vocabulary
\item Sentence order was randomized, and the data was split into 100 disjoint partitions
\item One such partition (1\%) of the data was chosen as the held-out set
\item The held-out set was then randomly shuffled and split again into 50 disjoint partitions to be used as development/test data
\item One such resulting partition (2\%, amounting to 159658 words without counting the sentence beginning marker \verb+<S>+ which is never predicted by the language model) of the held-out data were used as test data in our experiments; the remaining partitions are reserved for future experiments
\item The out-of-vocabulary (OoV) rate on the test set was 0.28\%.
\end{itemize}

The benchmark is available as a code.google.com project:
\url{https://code.google.com/p/1-billion-word-language-modeling-benchmark/}.
Besides the scripts needed to rebuild the training/held-out data, it also makes
available \href{https://drive.google.com/file/d/0B3u4EqGe3BUeMWhPS1hkdDZvTjA/edit?usp=sharing}{log-probability values} for each word in each of ten held-out data sets, for each of the baseline n-gram models.

Because the original data had already randomized sentence order, the benchmark is not
useful for experiments with models that capture long context dependencies across
sentence boundaries.


\section{Baseline Language Models}

As baselines we chose to use \cite{Katz:1987}, and Interpolated \cite{Kneser:1995} (KN) 5-gram LMs, as they are the most prevalent. Since in practice these models are pruned, often quite aggressivley, we also illustrate the negative effect of \cite{Stolcke:1998} entropy pruning on both models, similar to \cite{Chelba:2010}. In particular KN smoothing degrades much more rapidly than Katz, calling for a discerning choice in a given application.

\section{Advanced Language Modeling Techniques}

The number of advanced techniques for statistical language modeling is very large. It is out of scope of this paper to provide their detailed description, but we mention some of the most popular ones:
\begin{itemize}\addtolength{\itemsep}{-0.5\baselineskip}
\item N-grams with Modified Kneser-Ney smoothing~\cite{Chen:1996}
\item Cache~\cite{Jelinek:1991}
\item Class-based~\cite{Brown:1992}
\item Maximum entropy~\cite{Rosenfeld:1994}
\item Structured~\cite{Chelba:2000}
\item Neural net based~\cite{Bengio:2003}
\item Discriminative~\cite{Roark:2004}
\item Random forest~\cite{Xu:2005}
\item Bayesian~\cite{Teh:2006}
\end{itemize}
Below, we provide a short description of models that we used in our comparison using the benchmark data.

\subsection{Normalized Stupid Backoff}
The Stupid Backoff LM was proposed in~\cite{Brants:2007} as a simplified version of backoff LM,
suited to client-server architectures in a distributed computing environment.
It does not apply any discounting to relative frequencies, and it uses a single backoff weight instead of context-dependent backoff weights. As a result, the Stupid Backoff model does not generate normalized probabilities.
For the purpose of computing perplexity as reported in Table~\ref{results-table}, values output
by the model were normalized over the entire LM vocabulary.

\subsection{Binary Maximum Entropy Language Model}

The Binary MaxEnt model was proposed in~\cite{Xu:2011} and aims to avoid the expensive probability
normalization during training by using independent binary predictors. Each predictor
is trained using all the positive examples, but the negative examples are dramatically down-sampled.
This type of model is attractive for parallel training, thus we explored its performance further.

We trained two models with a sampling rate of 0.001 for negative examples, one uses n-gram features only
and the other uses n-gram and skip-1 n-gram features. We separated the phases of generating negative
examples and tuning model parameters such that the output of the first phase can be shared by two
models. The generation of the negative examples took 7.25 hours using
500 machines, while tuning the parameters using 5000 machines took 50 minutes, and 70 minutes for
the two models, respectively.

\begin{table*}
\begin{center}
\begin{tabular}{|l|c|c|c|c|}
\hline
\bf Model                           & \bf Perplexity \\
\hline
Interpolated KN 5-gram, 1.1B n-grams& 67.6           \\
\hline
\bf All models                      &\bf 43.8        \\
\hline
\end{tabular}
\end{center}
\caption{\label{combination-table}Model combination on the 1B Word Benchmark test set. The weights were tuned to minimize perplexity on held-out data. The optimal interpolation weights for the KN, rnn1024, rnn512, rnn256, SBO, HME were, respectively: 0.06, 0.61, 0.13, 0.00, 0.20, 0.00.}
\end{table*}

\subsection{Maximum Entropy Language Model with Hierarchical Softmax}

Another option to reduce training complexity of the MaxEnt models is to use a hierarchical
softmax~\cite{Goodman:2001b,Morin:2005}.
The idea is to estimate probabilities of groups of words, like in a class based model -- only the
classes that contain the positive examples need to be evaluated. In our case, we explored a
binary Huffman tree representation of the vocabulary, such that evaluation of
frequent words takes less time. The idea of using frequencies of words
for a hierarchical softmax was presented previously in~\cite{Mikolov:2011a}.

\subsection{Recurrent Neural Network Language Model}

The Recurrent Neural Network (RNN) based LM have recently achieved outstanding
performance on a number of tasks~\cite{Mikolov:2012}. It was shown that RNN LM significantly outperforms
many other language modeling techniques on the Penn Treebank data set~\cite{Mikolov:2011b}.
It was also shown that RNN models scale very well to data sets with hundreds of millions
of words~\cite{Mikolov:2011c}, although the reported training times for the largest models
were in the order of weeks.

We cut down training times by a factor of 20-50 for large problems using a
number of techniques, which allow RNN training in typically 1-10 days with
billions of words, $>1M$ vocabularies and up to 20B parameters on a single
standard machine without GPUs.

These techniques were in order of importance:
a) Parallelization of training across available CPU threads,
b) Making use of SIMD instructions where possible,
c) Reducing number of output parameters by 90\%,
d) Running a Maximum Entropy model in parallel to the RNN.
Because of space limitations we cannot describe the exact details of the speed-ups here -- they will be reported in an upcoming paper.

We trained several models with varying number of neurons (Table ~\ref{results-table}) using regular SGD with a learning rate of 0.05 to 0.001 using 10 iterations over the data. The MaxEnt models running in parallel to the RNN capture a history of 9 previous words, and the models use as additional features the previous 15 words independently of order. While training times approach 2 weeks for the most complex model, slightly worse models can be trained in a few days. Note that we didn't optimize for model size nor training speed, only test performance.

\section{Results}

\subsection{Performance of Individual Models}

Results achieved on the benchmark data with various types of LM are reported
in Table~\ref{results-table}.
We focused on minimizing the perplexity when choosing
hyper-parameters, however we also report the time required to train the models.
Training times are not necessarily comparable as they depend on the underlying implementation. Mapreduces can potentially process larger data sets than single-machine implementations, but come with a large overhead of communication and file I/O. Discussing details of the implementations is outside the scope as this paper.

\subsection{Model Combination}

The best perplexity results were achieved by linearly interpolating together probabilities from
all models. However, only some models had significant weight in the combination; the weights
were tuned on the held-out data. As can be seen in Table~\ref{combination-table}, the best
perplexity is about 35\% lower than the baseline - the modified Kneser-Ney smoothed 5-gram
model with no count cutoffs. This corresponds to about 10\% reduction of cross-entropy (bits).

Somewhat surprisingly the SBO model receives a relatively high weight in the linear combination of
models, despite its poor performance in perplexity, whereas the KN baseline receives a fairly small
weight relative to the other models in the combination.

\section{Conclusion}

We introduced a new data set for measuring research progress in statistical language modeling.
The benchmark data set is based on resources that are freely available on the web, thus fair comparison
of various techniques is easily possible. The importance of such effort is unquestionable: it
has been seen many times in the history of research that significant progress can be achieved when
various approaches are measurable, reproducible, and the barrier to entry is low.

The choice of approximately one billion words might seem somewhat restrictive. Indeed, it can be
hardly expected that new techniques will be immediately competitive on a large data set.
Computationally expensive techniques can still be compared using for example
just the first or the first 10 partitions of this new dataset, corresponding to approx.~10 million and 100 million words. However, to achieve impactful results
in domains such as speech recognition and machine translation, the language modeling
techniques need to be scaled up to large data sets.

Another contribution of this paper is the comparison of a few novel modeling approaches
when trained on a large data set. As far as we know, we were able to train the largest
recurrent neural network language model ever reported. The performance gain is very promising;
the perplexity reduction of 35\% is large enough to let us hope for significant improvements in
various applications.

In the future, we would like to encourage other researchers to participate in our efforts
to make language modeling research more transparent. This would greatly help to transfer
the latest discoveries into real-world applications. In the spirit of a benchmark our first
goal was to achieve the best possible test perplexities regardless of model sizes or training
time. However, this was a relatively limited collaborative effort, and some well known techniques
are still missing. We invite other researchers to complete the picture by evaluating new,
and well-known techniques  on this corpus. Ideally the benchmark would also contain ASR or SMT
lattices/N-best lists,  such that one can evaluate application specific performance as well.

\end{document}